%% file: arxiv-version.tex
\documentclass[10pt,twocolumn,letterpaper]{article}

\usepackage{cvpr}              

\input{preamble}

\usepackage[dvipsnames]{xcolor}
\definecolor{cvprblue}{rgb}{0.21,0.49,0.74}
\usepackage[pagebackref,breaklinks,colorlinks,citecolor=cvprblue]{hyperref}

\def\paperID{1} 
\def\confName{3DV\xspace}
\def\confYear{2026\xspace}

\title{Radially Distorted Homographies, Revisited}

\author{%
Mårten Wadenbäck\textsuperscript{a}
\quad
Marcus Valtonen Örnhag\textsuperscript{b}
\quad
Johan Edstedt\textsuperscript{a}
\\
{\normalsize \textsuperscript{a}Linköping University \quad \textsuperscript{b}Ericsson Research}
}

\begin{document}
\maketitle
\input{sec/0_abstract}
\input{sec/1_body}
\input{sec/2_acknowledgements}
\cleardoublepage
{
    \small
    \bibliographystyle{ieeenat_fullname}
    \bibliography{references_deprecated}
}

\end{document}

%% file: preamble.tex
\usepackage[british]{babel}
\usepackage[style=british]{csquotes}
\usepackage[all,british]{foreign}

\usepackage{amssymb}
\usepackage{amsthm}
\usepackage{mathtools}
\usepackage[all]{onlyamsmath}
\usepackage{shorthandbold}

\DeclareMathOperator{\adj}{adj}
\DeclareMathOperator{\diag}{diag}
\DeclarePairedDelimiterX{\norm}[1]{\lVert}{\rVert}{#1}
\DeclareRobustCommand{\T}{\top}
\DeclareRobustCommand{\bGamma}{\symbfup{\Gamma}}

\DeclareRobustCommand{\bXi}{\symbfup{\Xi}}
\DeclarePairedDelimiterX{\set}[1]{\lbrace}{\rbrace}{#1}

\usepackage[detect-all,per-mode=symbol]{siunitx}
\AtEndPreamble{\crefname{equation}{}{}}

\usepackage{tabularray}
\UseTblrLibrary{booktabs}
\UseTblrLibrary{siunitx}
 \usepackage{multirow}
\usepackage{tikz}

%% file: sec/0_abstract.tex
\begin{abstract}
    Homographies are among the most prevalent transformations occurring in geometric computer vision and projective geometry, and homography estimation is consequently a crucial step in a wide assortment of computer vision tasks.
    When working with real images, which are often afflicted with geometric distortions caused by the camera lens, it may be necessary to determine both the homography and the lens distortion---particularly the radial component, called \emph{radial distortion}---\emph{simultaneously} to obtain anything resembling useful estimates.
    When considering a homography with radial distortion between two images, there are three conceptually distinct configurations for the radial distortion; \begin{enumerate*}[label={(\roman*)}] \item distortion in only one image, \item identical distortion in the two images, and \item independent distortion in the two images. \end{enumerate*}
    While these cases have been addressed separately in the past, the present paper provides a novel and unified approach to solve all three cases.
    We demonstrate how the proposed approach can be used to construct new fast, stable, and accurate minimal solvers for radially distorted homographies.
    In all three cases, our proposed solvers are faster than the existing state-of-the-art solvers while maintaining similar accuracy. The solvers are tested on well-established benchmarks including images taken with fisheye cameras.
    A reference implementation of the proposed solvers is made available as part of HomLib\footnote{\url{https://github.com/marcusvaltonen/HomLib}}.
\end{abstract}

%% file: sec/1_body.tex
\section{Introduction}\label{sec:introduction}
Homographies constitute an important class of transformations that are often used in algorithms for solving computer vision tasks such as precalibration~\cite{cvpr/1999/sturm_maybank,pami/2000/zhang,pami/2007/hartley_kang}, autocalibration~\cite{eccv/1998/triggs}, stereo and multi-camera calibration~\cite{iccv/2003/ueshiba_tomita,ao/2005/hu_tan,3dv/2016/zhu_etal,ao/2017/guan_etal}, colour calibration~\cite{pami/2019/finlayson_etal}, metric rectification~\cite{cvpr/1998/liebowitz_zisserman,ijcv/2000/criminisi_etal}, stereo rectification~\cite{cvpr/1993/hartley_gupta,ijcv/1999/hartley,cvpr/1999/loop_zhang,prl/2010/kumar_etal}, ego-motion estimation~\cite{icra/2002/liang_pears,jfr/2015/zienkiewicz_davison,icip/2016/wadenback_etal,icpram-selected/2019/valtonen-ornhag_wadenback,icpram/2020/valtonen-ornhag,wacv/2021/valtonen-ornhag_etal}, panoramic stitching and mosaicing~\cite{cga/1996/szeliski,ijcv/2007/brown_lowe,cvpr/2007/brown_etal,jfr/2009/nicosevici_etal,cvpr/2011/gao_etal,cvpr/2013/zaragoza_etal}, scene understanding~\cite{cvprw/2006/perera_etal,its/2010/arrospide_etal,iccv/2011/flint_etal}, visual servoing~\cite{ijrr/2007/benhimane_malis,tac/2009/hu_etal}, augmented reality~\cite{isar/2000/simon_etal,isar/2000/seo_hong,cga/2002/prince_etal}, and more.

\begin{figure}[t!]
    \centering
    \begin{tikzpicture}
        \node[inner sep=0] (bg) at (0,0) {\includegraphics[width=\linewidth]{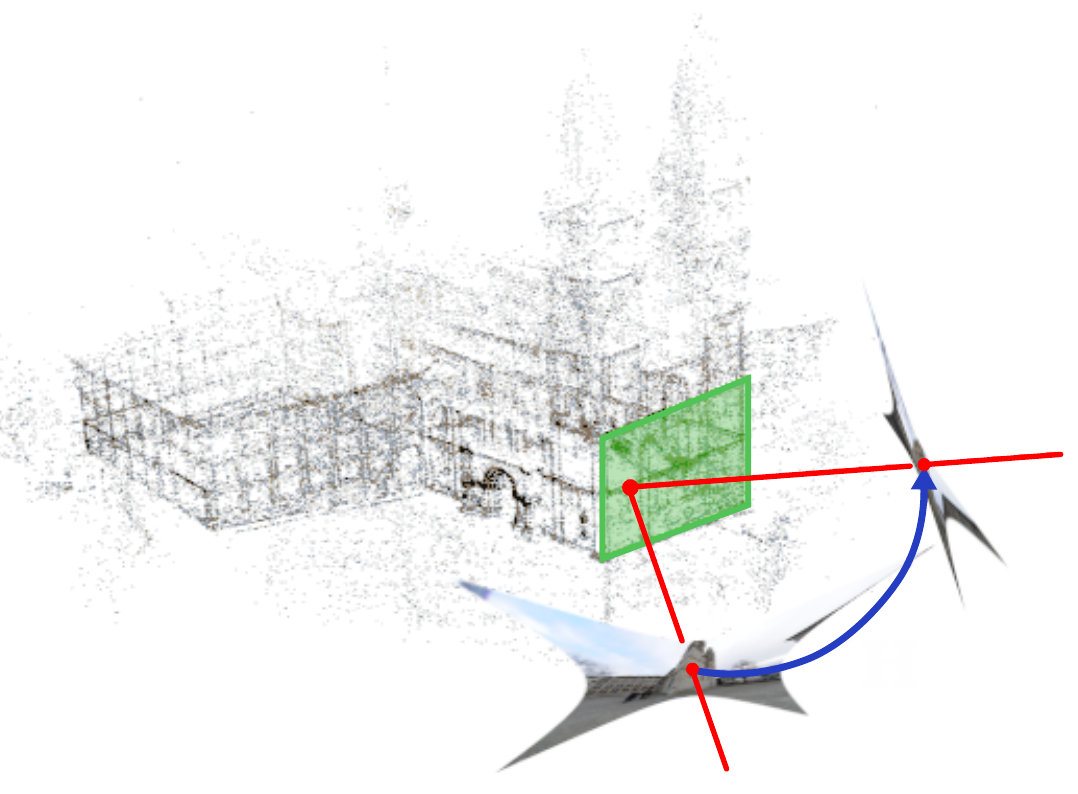}};
        \node at (2.8,-2) {\large $\bH$};
        \node at (1.1,-2.6) {\large $\lambda$};
        \node at (3.2,-0.1) {\large $\lambda'$};
    \end{tikzpicture}
    \caption{%
        We propose three novel homography solvers simultaneously estimating a homography between two views and radial distortion coefficients.
        It is well-known that a homography~$\bH$ maps image point correspondences which lie on a planar surface (\cf the green plane in the scene); however, physical cameras may deviate significantly from the pinhole camera model due to \eg radial distortion.
        In this figure of the \emph{Grossmünster church} the input images are significantly distorted due to the camera being equipped with a fisheye lens.
        The reference 3D reconstruction was obtained using RadialSfM~\cite{eccv/2020/larsson_etal}.
    }
    \label{fig:front}
\end{figure}
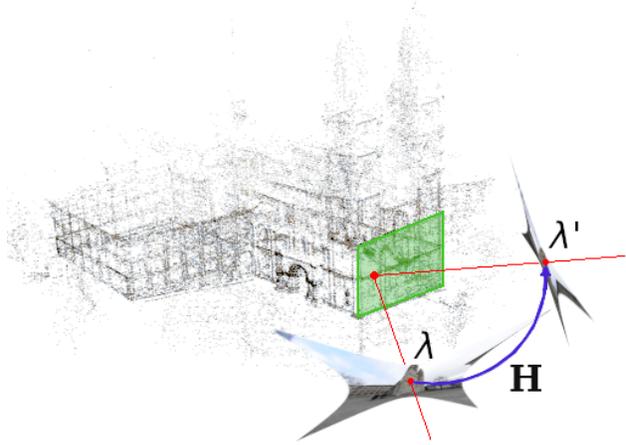

\begin{figure*}[t]
    \centering
    \includegraphics[width=\linewidth]{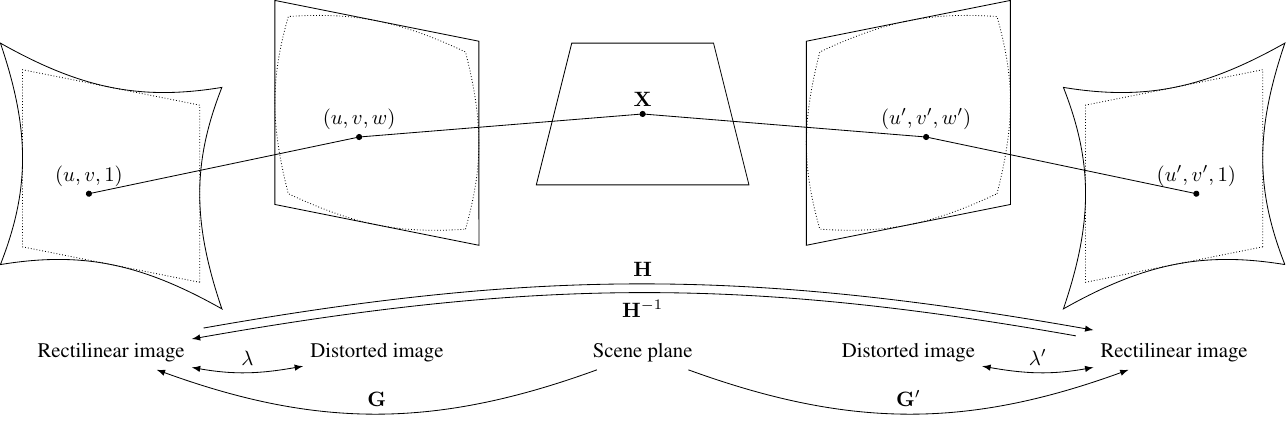}%
    \caption[dummy argument]{%
        To each of the rectilinear image planes, there is a homography ($\bG$ and $\bG'$, respectively) from the scene plane (which itself is inherently rectilinear).
        There is also a homography between the two rectified image planes.
        For the distortion, three distinct cases are therefore of interest here:
        \begin{enumerate*}[label={(\roman*)}] \item the one-sided case $\lambda'=0$, \item the two-sided equal case $\lambda'=\lambda$, and \item the two-sided independent case (where $\lambda$ and $\lambda'$ are independent). \end{enumerate*}
        The one-sided case is equivalent to just viewing the image plane with one camera.
    }
    \label{fig:problem geometry}
\end{figure*}

In most of these situations, the invocation of a homography is justified by the underlying geometry being considered, as a homography describes central projection of one projective plane onto another~\cite{book/2004/hartley_zisserman}.
For example, consider a perspective pinhole camera viewing a plane in the scene.
In this case, the mapping of homogeneous coordinates from the scene plane to the image plane will be a homography, \cf~\cite{pami/2000/zhang}.
Alternatively, considering two such views of a plane in the scene, the mapping between the two views will, again, be a homography, for the image region that contains the plane~\cite{book/2004/hartley_zisserman}.

Algebraically, a homography is represented by a (unique up to scale) non-singular matrix $\bH \in \mathbb{R}^{3\times 3}$ that maps homogeneous coordinates $\bx$ in one plane to $\bx'$ in the other plane according to $\bx' \sim \bH\bx$, where \enquote{$\sim$} denotes equality up to scale~\cite{book/2011/richter-gebert,book/2004/hartley_zisserman}.
To fixate the eight degrees of freedom of~$\bH$, it is sufficient to specify four point correspondences, \ie four points and their respective images under~$\bH$, as long as no three points (or their images under $\bH$) are collinear.
If this non-collinearity condition is met, the points are said to be in \emph{general position}.
Given at least four such correspondences, one can use the well-known \emph{Direct Linear Transformation} (DLT) to solve for $\bH$ linearly~\cite{book/2004/hartley_zisserman}.
Since the pinhole perspective camera does not model lenses, it is typically used together with a separate distortion model such as the Brown--Conrady model (see \eg~\cite{book/2022/szeliski}) or the one-parameter division model~\cite{dagm/1987/lenz,cvpr/2001/fitzgibbon,ipol/2014/aleman-flores_etal} when applied to real images.
For this reason, homography estimation methods based on DLT but which also consider radial distortion have been proposed in the literature~\cite{cvpr/2001/fitzgibbon,cvpr/2015/kukelova_etal,icpr/2024/nakano}.

Methods for automatically establishing correspondences generally cannot be guaranteed to only produce correct matches, so in practice the set of correspondences will contain outliers in the form of spurious correlations.
Such outliers have the potential to completely invalidate any subsequent model estimation, and it is important to reject them before proceeding.
This is often done using RANSAC~\cite{acm/1981/fischler_bolles} or one of its more refined incarnations~\cite{iccv/2003/nister,pami/2012/raguram_etal,cviu/2000/torr_zisserman,cvpr/2005/chum_matas,dagm/2003/chum_etal}, where a consensus set of inliers is found by repeatedly fitting the model to small random samples and then checking which of the other data also support this model.
The fewer data that are used for fitting the model in RANSAC, the higher is the chance of drawing an uncontaminated sample that only contains inliers, and thus hopefully in the process gains support from most of the other inliers.
For this reason, considerable effort has been directed to create so called \emph{minimal solvers} for different geometric estimation problems, \ie algorithms that use as few data as possible while ensuring a finite number of solutions.
Minimal solvers can provide both likely inlier sets and reasonable initialisations for further refinement such as bundle adjustment~\cite{iwva/1999/triggs_etal}.

\paragraph{Our contributions.}
Inspired primarily by earlier work by~\citet{cvpr/2015/kukelova_etal} and more recent work by~\citet{icpr/2024/nakano}, we revisit the important problem of estimating radially distorted homographies, and demonstrate that by replacing the DLT with a closed-form expression for $\bH$~\cite{book/1910/veblen_young,book/1962/seidenberg,book/2011/richter-gebert,prl/2019/guo}, we obtain numerically stable minimal solvers that are faster than the current state-of-the-art solvers for all possible configurations, \ie \begin{enumerate*}[label={(\roman*)}] \item the \emph{one-sided case}, \item the \emph{two-sided equal case}, and \item the \emph{two-sided independent case}. \end{enumerate*}
An illustration of the problem geometry for the three cases is shown in \cref{fig:problem geometry}.

\paragraph{Organisation of the paper.}
The rest of the paper is organised as follows.
In \cref{sec:related work} we discuss relevant related work, including current state-of-the-art methods for radially distorted homographies.
We formulate the problem mathematically and recount the classical closed-form formula for homographies in \cref{sec:problem formulation}.
Our proposed solvers are derived in \cref{sec:deriving the solvers}.
The solvers are evaluated in \cref{sec:experiments}, and \cref{sec:conclusion} concludes the paper.

\section{Related Work}\label{sec:related work}
As mentioned in the introduction, homographies are a crucial component of many camera calibration algorithms.
For example,~\citet{cvpr/1999/sturm_maybank} use homographies from scene planes to provide constraints on the intrinsic camera parameters.
Similarly,~\citet{pami/2000/zhang} computes homographies from a planar calibration target to the image plane using DLT, but in contrast to~\cite{cvpr/1999/sturm_maybank} also incorporates radial distortion in a final non-linear refinement step, which forms the basis for the widely used calibration method in OpenCV.

The first work to include radial distortion already at the homography computation step is due to \citet{cvpr/2001/fitzgibbon}, who extended the DLT equations into a quadratic eigenvalue problem to capture lens distortion modelled by the one-parameter division model~\cite{dagm/1987/lenz,cvpr/2001/fitzgibbon}.

Subsequently, solvers that make use of additional assumptions have been proposed for some notable special cases, \eg for panoramic stitching~\cite{cvpr/2007/brown_etal,cvpr/2008/jin,bmvc/2009/byrod_etal} where images are assumed to be captured with coinciding camera centres.
Instead of eliminating the translation, another option is to assume that the rotation is known, which can make sense in drone applications where an IMU can provide sufficiently accurate rotation estimates~\cite{wacv/2021/valtonen-ornhag_etal}.
The present paper, however, does not make any such simplifying assumptions for the scene or pose parameters.

We consider three different cases for radially distorted homographies, which we will refer to as the \emph{one-sided case}, the \emph{two-sided equal case}, and the \emph{two-sided independent case} (see \cref{fig:problem geometry}).
Each of these cases has been considered separately in previous work, but we take a novel unified approach for all three cases.
Starting chronologically, \citet{cvpr/2001/fitzgibbon} solved the \emph{two-sided equal case } by deriving a quadratic eigenvalue problem with up to 18 real solutions.
\citet{cvpr/2015/kukelova_etal} proposed two solvers for the \emph{two-sided independent case}, \ie for radially distorted homographies with independent distortion coefficients $\lambda$ and $\lambda'$.
Their minimal 5 point solver uses a Gröbner basis approach, and requires performing Gauss--Jordan elimination on a \numproduct{16 x 21} template matrix.
Their non-minimal approach uses 6 points, and is computationally significantly cheaper, but while less sensitive to noise is more sensitive to outliers due to the inclusion of the additional point.
Recently, \citet{icpr/2024/nakano} proposed a DLT-based solution to the \emph{one-sided case} $\lambda' = 0$, finding the parameters of the transformation from the null-space to a \numproduct{10 x 12} design matrix.
For an overview comparison of some of the key properties (including timings) of the existing solvers and our proposed solvers, see \cref{tab:timings}.

\section{Problem formulation}\label{sec:problem formulation}
Let us start by considering the problem without radial distortion.
Given a number of point correspondences, $\bx_j' \leftrightarrow \bx_j$ for $j=1,\ldots,n$, expressed using homogeneous coordinates $\bx_j=(u_j,v_j,w_j)$ and $\bx_j'=(u_j',v_j',w_j')$, we want to find a non-singular $\bH \in \mathbb{R}^{3\times 3}$ such that
\begin{equation}
    \alpha_j\bx_j'
    = \bH\bx_j,
    \quad
    \alpha_j\in\mathbb{R},
    \quad
    j
    = 1,\ldots,n.
\end{equation}
In the computer vision literature, this problem is traditionally solved using the \emph{Direct Linear Transformation} (DLT) \cite{book/2004/hartley_zisserman,book/2022/szeliski}, and this route was used as a foundation for the solvers with radial distortion introduced in \cite{cvpr/2001/fitzgibbon}, \cite{cvpr/2015/kukelova_etal} and \cite{icpr/2024/nakano}.
We will instead use a much older approach, which we shall discuss next, as a springboard to constructing our proposed solvers.

\subsection{The classical closed-form solution}
While largely overlooked by the computer vision community, the classical closed-form solution for a homography has a long history and is well established in the literature~\citep{book/1910/veblen_young,book/1962/seidenberg,book/2011/richter-gebert}.
Its core idea is the fact that it is easy to find the mapping to (or from) the four intermediary points $\set{(1,0,0),\,(0,1,0),\,(0,0,1),\,(1,1,1)}$. 
More specifically, it works as follows.
Let $n=4$ and assume that the points are in general position.
This means that we can find non-zero scalars $\gamma_1,\gamma_2,\gamma_3$ such that
\begin{equation}
    \bx_4
    = \gamma_1\bx_1+\gamma_2\bx_2+\gamma_3\bx_3,
\end{equation}
and likewise, non-zero scalars $\gamma_1',\gamma_2',\gamma_3'$ such that
\begin{equation}
    \bx_4'
    = \gamma_1'\bx_1'+\gamma_2'\bx_2'+\gamma_3'\bx_3'.
\end{equation}
Indeed, if one of these scalars were to be zero, that would contravene our assumption that the points are in general position, \ie, three of the points would need to be collinear on at least one side of the transformation.

Now, the homography $\bU = \begin{pmatrix} \gamma_1\bx_1 & \gamma_2\bx_2 & \gamma_3\bx_3 \end{pmatrix}$ maps the intermediary points to $\bx_1,\ldots,\bx_4$, and in the same way, $\bU' = \begin{pmatrix} \gamma_1'\bx_1' & \gamma_2'\bx_2' & \gamma_3'\bx_3' \end{pmatrix}$ maps the intermediary points to $\bx_1',\ldots,\bx_4'$.
Hence, the sought homography is obtained by reversing the first mapping and applying the mappings in succession, yielding the composite mapping $\bH = \bU'\bU^{-1}$.

\begin{figure*}[t]
    \centering
    \includegraphics[width=0.48\linewidth]{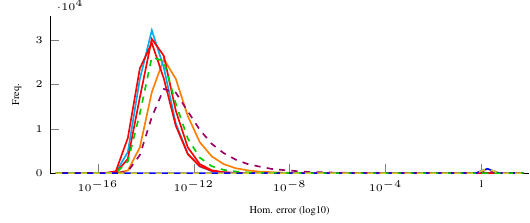}%
    \includegraphics[width=0.48\linewidth]{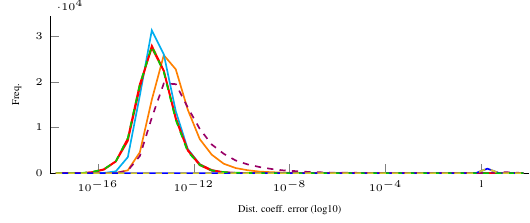}\\%
    \includegraphics[width=0.65\linewidth]{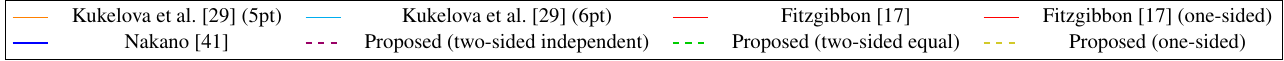}%
    \caption{\emph{Numerical stability}. Homography errors (left) and distortion coefficient error (right) for \num{10000} noise-free randomly generated problem instances.}
    \label{fig:stability}
\end{figure*}

We may write this procedure more compactly by denoting $\bGamma = (\gamma_1,\gamma_2,\gamma_3)$ and $\bGamma' = (\gamma_1',\gamma_2',\gamma_3')$, as well as
\begin{equation}
    \label{eq:xis}
    \bXi
    =
    \begin{pmatrix}
        \bx_1 & \bx_2 & \bx_3 \\
    \end{pmatrix}
    \quad\textup{and}\quad
    \bXi'
    =
    \begin{pmatrix}
        \bx_1' & \bx_2' & \bx_3' \\
    \end{pmatrix}
    .
\end{equation}
With this notation, we now have
\begin{equation}
    \label{eq:gammas}
    \bGamma
    =
    \bXi^{-1}
    \bx_4
    \quad\textup{and}\quad
    \bGamma'
    =
    (\bXi')^{-1}
    \bx_4'.
\end{equation}
Then, finally, $\bH$ can be computed in closed-form as
\begin{equation}
    \label{eq:direct H}
    \bH
    = \bXi' (\diag{\bGamma'}) (\diag{\bGamma})^{-1} \bXi^{-1}.
\end{equation}
There are some small and obvious improvements that we shall wish to apply to the basic method described above.
First of all, since
\begin{equation}
    \bM\adj{\bM}
    = \det(\bM)\bI
\end{equation}
for any square matrix $\bM$~\cite{book/1944/aitken}, the non-diagonal inverses in \cref{eq:gammas,eq:direct H} can be replaced with adjugates, as the scalar factor is irrelevant when working with homogeneous coordinates.
Recalling the explicit formula for \numproduct{3 x 3} adjugates (see \eg Ch.~3 in \cite{book/1944/aitken}), we may write
\begin{equation}
    \adj{\bXi}
    = \!
    \begin{pmatrix}
        v_2w_3{-}v_3w_2\!\! & \!u_3w_2{-}u_2w_3\!\! & \!u_2v_3{-}u_3v_2 \\
        v_3w_1{-}v_1w_3\!\! & \!u_1w_3{-}u_3w_1\!\! & \!u_3v_1{-}u_1v_3 \\
        v_1w_2{-}v_2w_1\!\! & \!u_2w_1{-}u_1w_2\!\! & \!u_1v_2{-}u_2v_1 \\
    \end{pmatrix}
    \!,
\end{equation}
and similarly for $\adj(\bXi')$.
These are now plugged into \cref{eq:gammas,eq:direct H} to give
\begin{equation}
    \label{eq:new gammas}
    \bGamma
    =
    \adj(\bXi)
    \bx_4,
    \quad
    \bGamma'
    =
    \adj(\bXi')
    \bx_4',
\end{equation}
and
\begin{equation}
    \label{eq:new direct H}
    \bH
    = \bXi' (\diag{\bGamma'}) (\diag{\bGamma})^{-1} \adj(\bXi).
\end{equation}
Secondly, if all points are finite and normalised to have their last entry equal to one, the resulting expression \cref{eq:new direct H} becomes a little simpler and requires fewer arithmetic operations to evaluate \cite{prl/2019/guo}.

\subsection{Including radial distortion}\label{sec:including radial distortion}
Assuming the standard one-parameter division model for radial distortion \cite{dagm/1987/lenz,cvpr/2001/fitzgibbon}, we have
\begin{equation}
    \begin{aligned}
        w_j
        &= 1+\lambda(u_j^2+v_j^2), \quad \textup{and}\\
        w_j'
        &= 1+\lambda'\big((u_j')^2+(v_j')^2\big).\\
    \end{aligned}
\end{equation}
Three distinct cases are of interest here (\cf~\cref{fig:problem geometry}):
\begin{enumerate*}[label={(\roman*)}] \item the one-sided case $\lambda'=0$, \item the two-sided equal case $\lambda'=\lambda$, and \item the two-sided independent case (where $\lambda$ and $\lambda'$ are independent). \end{enumerate*}

\section{Deriving the solvers}\label{sec:deriving the solvers}
For the sake of generality, let us initially assume that $\lambda$ and $\lambda'$ are independent and potentially both non-zero.
Then, for a fifth point correspondence $\bx_5' \leftrightarrow \bx_5$, we can use \cref{eq:new direct H} to write
\begin{equation}
    \label{eq:fifth correspondence}
    \adj(\diag{\bGamma'}) \adj(\bXi') \bx_5'
    \sim
    \adj(\diag{\bGamma}) \adj(\bXi)\bx_5,
\end{equation}
where the left hand side is a vector valued cubic polynomial $\bN'(\lambda') \in \mathbb{R}^{3}[\lambda']$ and the right hand side will be a cubic polynomial $\bN(\lambda) \in \mathbb{R}^3[\lambda]$ (the expressions for the coefficients will be identical up to toggling of the primes).

Using the explicit formula for the adjugate of a \numproduct{3 x 3} matrix again, we can expand both sides as necessary and read off the coefficients of $\bN(\lambda)$ and $\bN'(\lambda')$.

\subsection{The one-sided case $\lambda' = 0$}
This is the case considered by~\citet{icpr/2024/nakano}, and it arises naturally in connection with camera calibration.
Here, the goal is to find the homography from the scene plane (\eg a planar calibration target) to the rectified scene plane, together with the distortion coefficient $\lambda$.
An alternative, equivalent interpretation, is that of finding a homography between a distorted and a rectified image.

Setting $\lambda' = 0$ in \cref{eq:fifth correspondence} and taking the cross product yields
\begin{equation}
    \bN(\lambda)
    \sim \bN'(0)
    \iff
    \bN(\lambda) \times \bN'(0)
    = \bzero,
\end{equation}
which corresponds to three scalar cubic equations in $\lambda$, which will have up to three real-valued solutions.
We can solve any one of them using standard techniques, \eg Cardano's formula~\cite{book/1944/turnbull}, the trigonometric method~\cite{book/1944/turnbull}, the companion matrix method~\cite{sjmaa/2015/aurentz_etal} or Sturm sequences~\cite{book/1944/turnbull}.
For its speed, and to avoid complex numbers, we use the trigonometric method.

\begin{figure*}[t]
    \centering
    \includegraphics[width=0.33\linewidth]{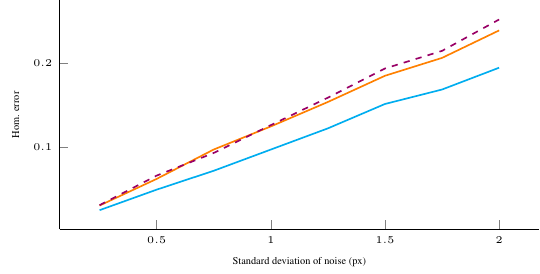}%
    \includegraphics[width=0.33\linewidth]{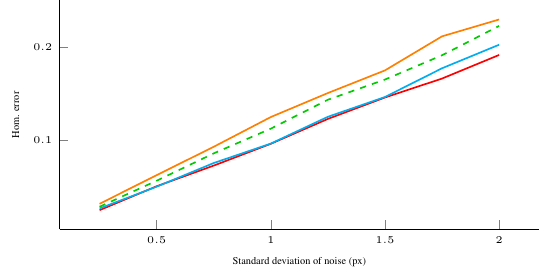}%
    \includegraphics[width=0.33\linewidth]{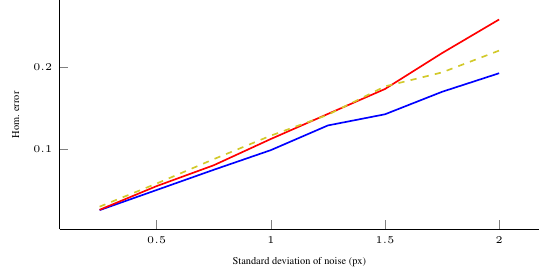}\\
    \includegraphics[width=0.33\linewidth]{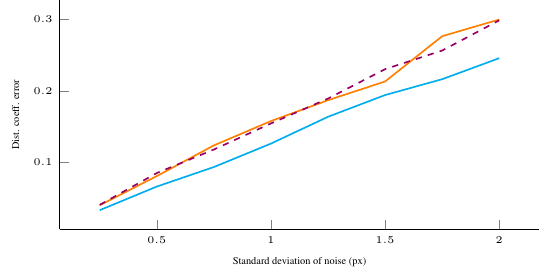}%
    \includegraphics[width=0.33\linewidth]{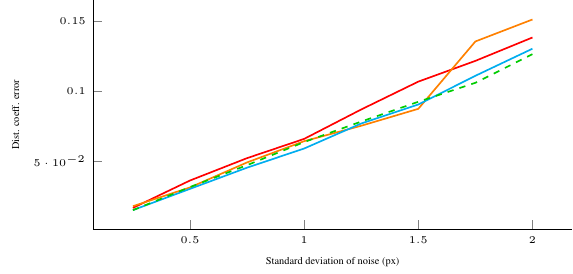}%
    \includegraphics[width=0.33\linewidth]{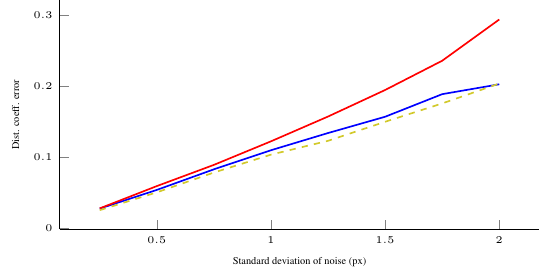}\\%
    \includegraphics[width=0.65\linewidth]{plots/legend.pdf}%
    \caption{%
        \emph{Sensitivity to noise}.
        Homography errors (top) and distortion coefficient errors (bottom) for different noise levels.
        Each point reports the median error of \num{10000} randomly generated problem instances (per noise level).
        The left column shows the noise sensitivity for the two-sided independent solvers, the middle column shows the noise sensitivity for the two-sided equal solvers, and the right column shows the noise sensitivity for the one-sided solvers.
    }
    \label{fig:noise}
\end{figure*}

\subsection{The two-sided equal case}
This is the case originally considered by~\citet{cvpr/2001/fitzgibbon}, and arises naturally in situations with two images taken by the same camera (and without changing the optical configuration), for example when taking shots for a panorama.
Setting $\lambda' = \lambda$ in \cref{eq:fifth correspondence} yields
\begin{equation}
    \bN(\lambda)
    \sim \bN'(\lambda)
    \iff
    \bN(\lambda) \times \bN'(\lambda)
    = \bzero,
\end{equation}
which corresponds to three scalar sextic equations in $\lambda$.
We can again solve any one of them using a standard technique of our choice, and it will result in up to six real-valued solutions.
To avoid computations involving complex numbers, we use Sturm sequences~\cite{book/1944/turnbull}.

\subsection{The two-sided independent case}
This is the case considered by~\citet{cvpr/2015/kukelova_etal}, and arises naturally \eg when stitching images coming from different cameras (or if the optical configuration has changed, \eg due to zoom or focus adjustments).
From \cref{eq:fifth correspondence} we get
\begin{equation}
    \bN(\lambda)
    \sim \bN'(\lambda')
    \iff
    \bN(\lambda) \times \bN'(\lambda')
    = \bzero,
\end{equation}
which corresponds to three scalar polynomial equations of degree six in $\lambda$ and $\lambda'$.

A polynomial system of equations of this type can be solved \eg using resultant-based methods or Gröbner basis methods.
We use the automatic Gröbner basis generator proposed in \cite{cvpr/2017/larsson_etal} to create a numerical solver.
The system has in general nine complex-valued solutions.
Attentive readers note that the solver by \citet{cvpr/2015/kukelova_etal} only had five potential solutions---the four extra solutions obtained from our solver are introduced due to the adjugate matrix computations and can therefore be discarded by checking the determinant of the corresponding matrix.
By using the basis heuristic proposed in~\cite{cvpr/2018/larsson_etal}, an elimination template of size \numproduct{9 x 18} was found, which combined with Sturm sequence root finding generates a numerically stable and computationally efficient solver.

\section{Experiments}\label{sec:experiments}
We compare our solvers against the existing state-of-the-art solvers, \ie \citet{icpr/2024/nakano} for the one-sided case, \citet{cvpr/2001/fitzgibbon} for the two-sided equal case, and \citet{cvpr/2015/kukelova_etal} for the two-sided independent case.
All solvers are implemented in C++ using Eigen, and timings are shown in~\cref{tab:timings}.
The results show the median value of \num{100000} random problem instances, executed on a standard laptop equipped with an 11th Gen Intel\textsuperscript{\small\textregistered} Core\textsuperscript{\small\texttrademark} i5-1145G7 @\qty{2.6}{\giga\hertz} CPU.
While this reporting of the runtime gives some general sense of the speed of the solver, additional aspects, \eg the sample size and the typical number of solutions, will influence the effective runtime.
\begin{table}[ht!]
    \centering
    \caption{Properties of the evaluated state-of-the-art solvers including the median runtime for \num{100000} random problem instances.}
    \begin{tblr}{colspec={XQ[r]Q[r]Q[r]Q[r]},rowsep=0ex,width=\linewidth}
        \toprule
        Solver & Pts. & Sols. & Min. & Time \\
        \midrule
        \SetRow{ht=1.1\baselineskip}\SetCell[c=5]{c} Two-sided with independent $\lambda$ and $\lambda'$ \\
        \midrule
        \citet{cvpr/2015/kukelova_etal} & 5 & 5 & $\checkmark$ & \qty{34.8}{\micro\second} \\
        \citet{cvpr/2015/kukelova_etal} & 6 & 2 & & \qty{12.3}{\micro\second} \\
        Proposed & 5 & 5 & $\checkmark$ & \textbf{\qty{8.1}{\micro\second}} \\
        \midrule
        \SetRow{ht=1.1\baselineskip}\SetCell[c=5]{c} Two-sided with $\lambda' = \lambda$ \\
        \midrule
        \citet{cvpr/2001/fitzgibbon} & 5 & 18 & & \qty{64.0}{\micro\second} \\
        \citet{cvpr/2015/kukelova_etal} & 5${}^*$ & 5 & & \qty{25.6}{\micro\second} \\
        \citet{cvpr/2015/kukelova_etal} & 6${}^*$ & 2 & & \qty{11.2}{\micro\second} \\
        Proposed & 4.5 & 6 & $\checkmark$ & \textbf{\qty{3.3}{\micro\second}} \\
        \midrule
        \SetRow{ht=1.1\baselineskip}\SetCell[c=5]{c} One-sided, \ie $\lambda' = 0$ \\
        \midrule
        \citet{cvpr/2001/fitzgibbon} & 5${}^{**}$ & 6 & & \qty{17.6}{\micro\second} \\
        \citet{icpr/2024/nakano} & 4.5 & 3 & $\checkmark$ & \qty{4.5}{\micro\second} \\
        Proposed & 4.5 & 3 & $\checkmark$ & \textbf{\qty{0.6}{\micro\second}} \\
        \bottomrule
    \end{tblr}\\[2mm]
    \begin{minipage}{0.9\linewidth}
        \scriptsize
        ${}^*$: The solver is terminated early if distortion coefficients are not approximately equal, and the geometric mean is returned in case they are.\\
        ${}^{**}$: Only the two-sided case was originally treated by~\citet{cvpr/2001/fitzgibbon}; however, \citet{icpr/2024/nakano} showed how to generalise the approach to the one-sided case.
    \end{minipage}
    \label{tab:timings}
\end{table}

\begin{figure*}[t!]
    \centering
    \includegraphics[width=0.32\linewidth]{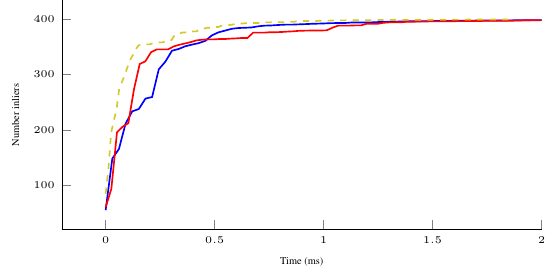}%
    \includegraphics[width=0.32\linewidth]{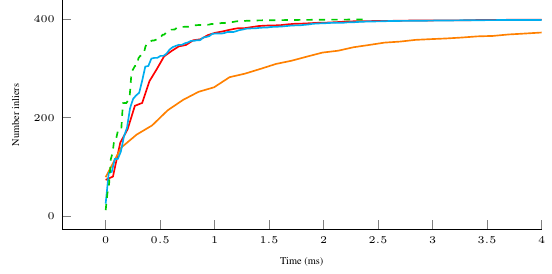}%
    \includegraphics[width=0.32\linewidth]{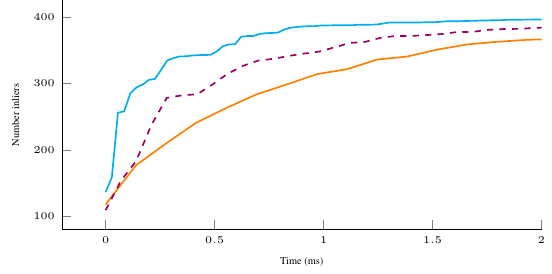} \\%
    \includegraphics[width=0.32\linewidth]{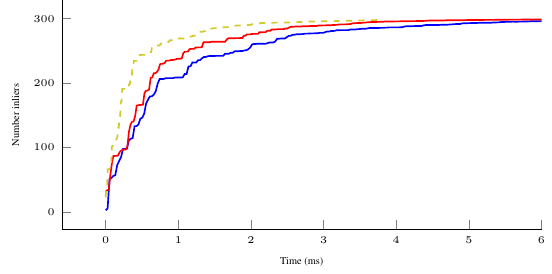}%
    \includegraphics[width=0.32\linewidth]{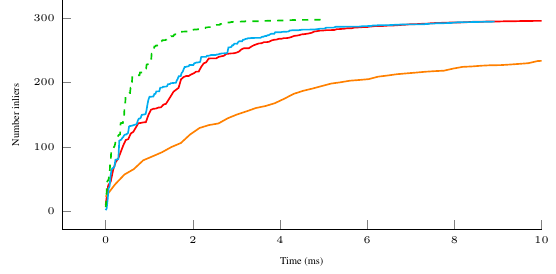}%
    \includegraphics[width=0.32\linewidth]{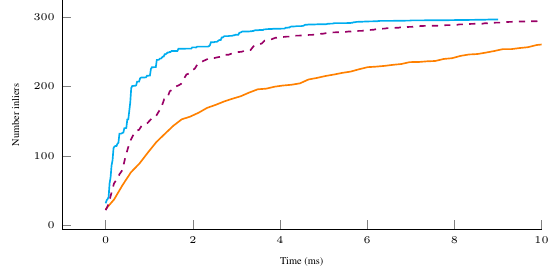} \\%
    \includegraphics[width=0.32\linewidth]{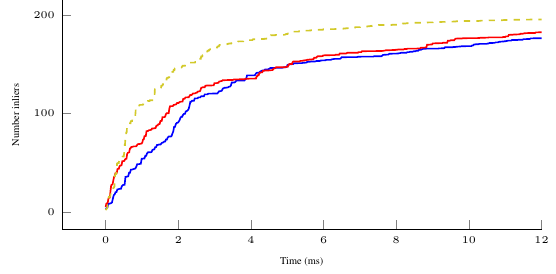}%
    \includegraphics[width=0.32\linewidth]{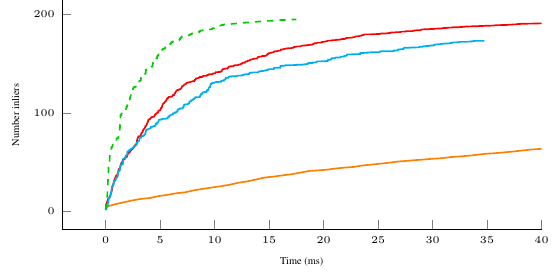}%
    \includegraphics[width=0.32\linewidth]{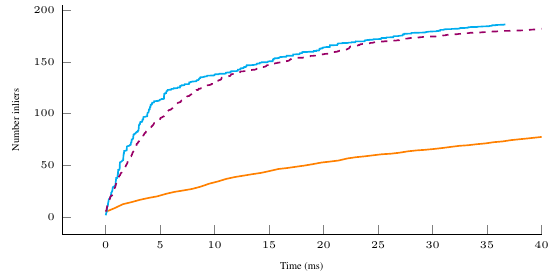} \\%
    \includegraphics[width=0.65\linewidth]{plots/legend.pdf}%
    \caption{%
        \emph{RANSAC integration}.
        Here we plot the cumulative number of inliers versus time for all the evaluated solver, for \qty{20}{\percent} outliers (top row), \qty{40}{\percent} outliers (mid row), and \qty{60}{\percent} outliers (bottom row).
        The left column shows the results for the one-sided solvers, the middle column shows the results for the two-sided equal solvers, and the right column shows the results for the two-sided independent solvers.
    }
    \label{fig:ransac}
\end{figure*}

\subsection{Numerical stability}
To evaluate the numerical stability of the solvers, we have generated \num{10000} scenes with 3D points distributed randomly on an unknown plane.
A pair of cameras with a depth of 0.1--10 to the scene plane, focal length = 1000, and distortion coefficients in the range $[-0.2,-0.01]$ were generated such that the field-of-view for both cameras is 70 degrees and the scene points are in front of the cameras.
This captures a low to medium level of distortion, for which the one-parameter division model is known to be a good approximation for typical physical cameras.

As can be seen in~\cref{fig:stability}, all the considered methods are numerically stable.
For the solvers with two distortion coefficients, the algebraic error is computed as the geometric mean of the individual ones, \ie $k_{\text{err}} = \sqrt{k_{1,\text{err}}k_{2,\text{err}}}$.

\subsection{Sensitivity to noise}
In addition to the numerical stability, we want to investigate the sensitivity of the proposed solvers with respect to Gaussian noise affecting the data.
Gaussian noise of varying standard deviation is therefore applied to the distorted (normalised) point correspondences and the corresponding median errors over \num{10000} random problem instances are reported, see~\cref{fig:noise}.

Overall, the evaluated solvers all have comparable noise sensitivity to the other solvers for the same case.
The strongest deviations are seen for \citet{cvpr/2001/fitzgibbon} which is somewhat more sensitive with respect to the distortion coefficient in the one-sided case, and for the 6 point version of \citet{cvpr/2015/kukelova_etal} in the two-sided independent case, which is less sensitive (this is somewhat expected, as this method uses an additional point, \cf~\cite{pami/2014/pham_etal}).
In situations with both high noise and a high inlier ratio, this could be advantageous, but conversely, if the noise and inlier ratio are low, the inclusion of the extra point may instead be an encumbrance.

\begin{figure*}[t!]
    \centering
    \includegraphics[width=0.32\linewidth]{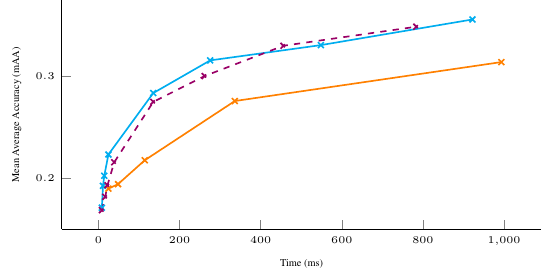}%
    \includegraphics[width=0.32\linewidth]{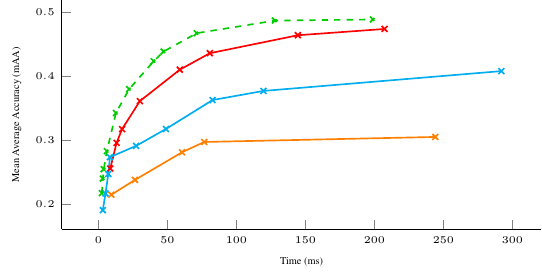}%
    \includegraphics[width=0.32\linewidth]{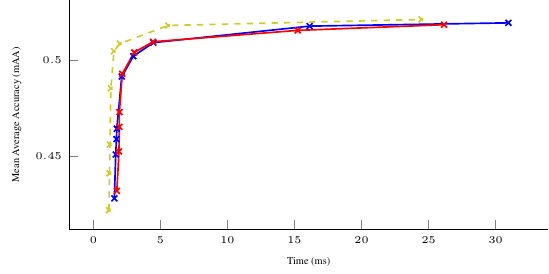}
    \includegraphics[width=0.65\linewidth]{plots/legend.pdf}%
    \caption{\emph{Results on HPatches}. The left column shows the results for one-sided solvers, the middle column shows the results for the two-sided equal solvers, and the right column shows the results for the two-sided independent solvers.}
    \label{fig:hpatches_time_mAA}
\end{figure*}

\subsection{Integration in a robust framework}
We integrate the polynomial solvers under evaluation into a RANSAC framework to get a better understanding of the performance.
While all solvers are fast, RANSAC also requires evaluating a hypothesis for each model estimated, which affects the total time.
We study the cumulative number of of inliers found versus execution time in~\cref{fig:ransac}, where each measurement shows the average over \num{1000} random problem instances.
Settings in previous sections apply, and the standard deviation of the noise is set to \qty{0.5}{px}, which is realistic for traditional descriptors such as SIFT~\cite{ijcv/2004/lowe}, and an inlier threshold of \qty{5}{px} is used.
We find that our proposed solvers for the one-sided and two-sided equal cases consistently find more inliers for different outlier ratios.
For the two-sided independent case, the 6 point solver by \citet{cvpr/2015/kukelova_etal} has a clear advantage due to it only having two solutions to evaluate, even though the margin decreases for higher outlier ratios.


\subsection{Evaluation on real data}
Next, we evaluate the proposed solvers on real images by
integrating them in a LOMSAC-framework~\cite{bmvc/2012/lebeda_etal} applying
a local optimisation step to the so-far-the-best model and non-linear refinement.

\begin{table}[ht!]
\centering
\caption{%
    \emph{Results on HPatches}.
    mAA (AUC score) when allowing the solvers to run up to \qty{1000}{\milli\second} (two-sided independent), \qty{250}{\milli\second} (two-sided equal), \qty{30}{\milli\second} (one-sided).
    Best results are marked in bold.
}
\label{tab:hpatches}
    \begin{tblr}{colspec={XQ[r]Q[r]Q[r]},rowsep=0ex,width=\linewidth}
        \toprule
        \SetCell[r=2]{m} Solver &  \SetCell[c=3]{c} mAA (AUC) \\
        \cline{2-4}
                                &  @\qty{1}{px}   & @\qty{3}{px}   & @\qty{5}{px}  \\
        \midrule
        \SetRow{ht=1.1\baselineskip}\SetCell[c=4]{c} Two-sided with independent $\lambda$ and $\lambda'$ \\
        \midrule
        \citet{cvpr/2015/kukelova_etal} (5pt)       & 0.0983 & 0.3343 & 0.5079 \\
        \citet{cvpr/2015/kukelova_etal} (6pt)       & \textbf{0.1704} & 0.3693 & 0.5276 \\
        Proposed (5pt)                              & 0.1261 & \textbf{0.3820} & \textbf{0.5379} \\
        \midrule
        \SetRow{ht=1.1\baselineskip}\SetCell[c=4]{c} Two-sided with $\lambda' = \lambda$ \\
        \midrule
        \citet{cvpr/2001/fitzgibbon} (5pt)          & 0.3092 & 0.5133 & 0.6310 \\
        \citet{cvpr/2015/kukelova_etal} (5pt)${}^*$ & 0.1127 & 0.3756 & 0.5379 \\
        \citet{cvpr/2015/kukelova_etal} (6pt)${}^*$ & 0.2186 & 0.4415 & 0.5867 \\
        Proposed (4.5pt)                            & \textbf{0.3227} & \textbf{0.5192} & \textbf{0.6329} \\
        \midrule
        \SetRow{ht=1.1\baselineskip}\SetCell[c=4]{c} One-sided, \ie $\lambda' = 0$ \\
        \midrule
        \citet{cvpr/2001/fitzgibbon} (5pt)${}^{**}$ & 0.2542 & 0.5996 & 0.7285 \\
        \citet{icpr/2024/nakano} (4.5pt)            & 0.2537 & 0.5992 & 0.7269 \\
        Proposed (4.5pt)                            & \textbf{0.2566} & \textbf{0.6019} & \textbf{0.7298} \\
        \bottomrule
    \end{tblr}
\end{table}

\paragraph{Evaluation on HPatches.}
For homography estimation, it is common to evaluate on the HPatches dataset \cite{cvpr/2017/balntas_etal}, is a Homography benchmark consisting of \num{116} planar scenes and comes with an accurate reference in the form of manually annotated correspondences that are often used as ground truth.
Unfortunately, HPatches does not have noticeable radial distortion, and the dataset does not contain a reference estimate for this.
As a sanity check, we nonetheless report the mAA (AUC scores) on HPatches for thresholds of \qty{1}{px}, \qty{3}{px} and \qty{5}{px} in \cref{tab:hpatches}.

We follow the evaluation settings in LightGlue~\cite{iccv/2023/lindenberger_etal} and resize the shorter side of the image to $480$ pixels, and use a total of $1024$ keypoints with SuperPoint~\cite{cvprw/2018/detone_etal}.
SuperPoint + LightGlue is used for establishing the correspondences.

Detailed scores for the maximum runtime is shown in~\cref{tab:hpatches}. 
Here we allowed the solvers to run up to \qty{1}{\second} (two-sided independent), \qty{250}{\milli\second} (two-sided equal), and \qty{30}{\milli\second} (one-sided), respectively.
mAA (AUC score) as a function of runtime is shown in~\cref{fig:hpatches_time_mAA}.
In general our proposed solvers yield higher mAA than previous solvers.

\begin{figure}[ht!]
    \centering
    \includegraphics[width=\linewidth]{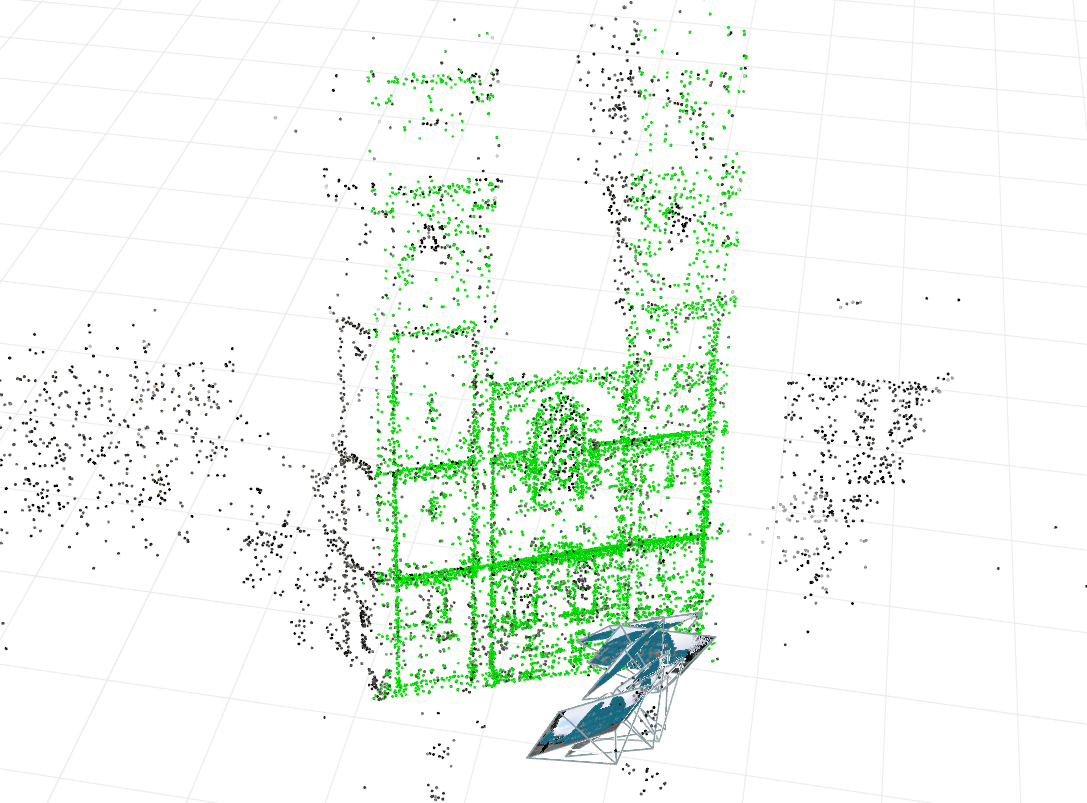}
    \caption{\emph{Grossmünster Church}. 3D reconstruction containing the planar points (green) and the seven camera positions.}
    \label{fig:grossmunster_gt}
\end{figure}

\begin{figure*}[t!]
    \centering
    \includegraphics[width=0.81\linewidth]{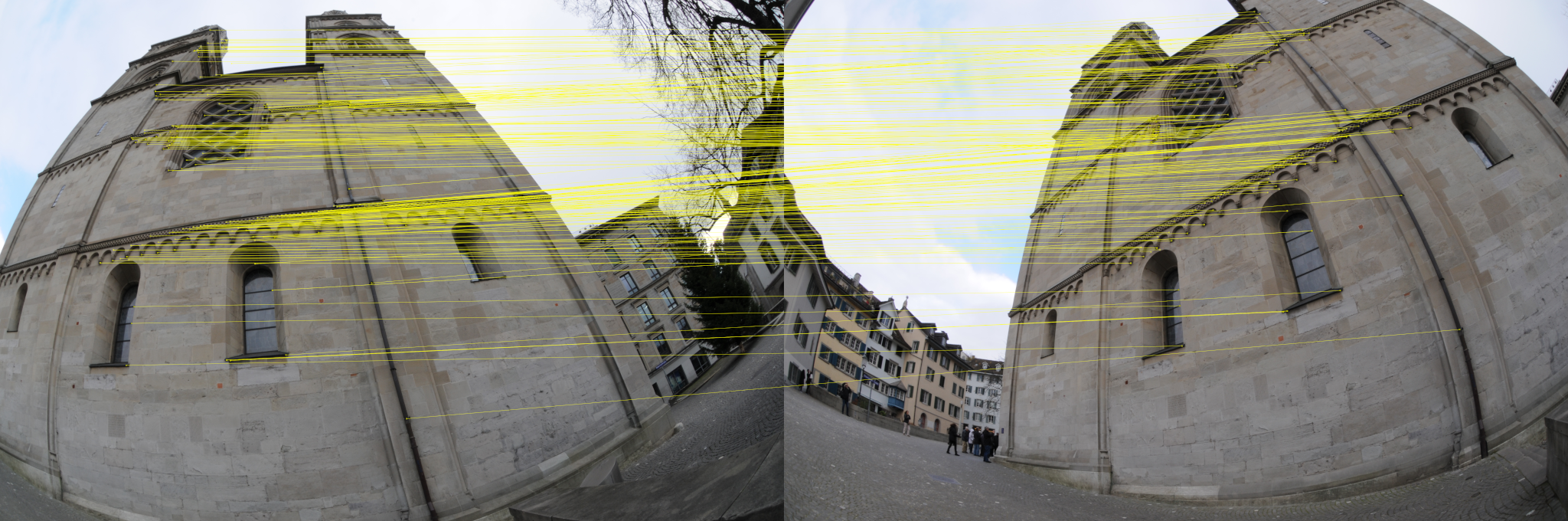}\\
    \includegraphics[width=0.27\linewidth]{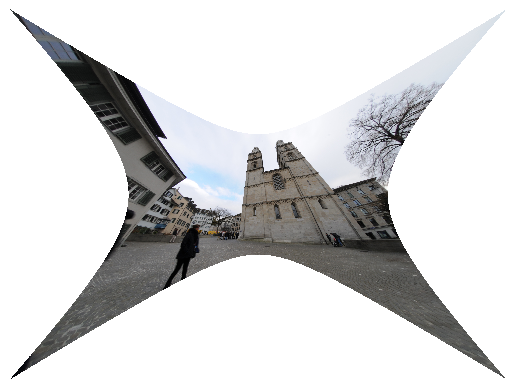}%
    \includegraphics[width=0.27\linewidth]{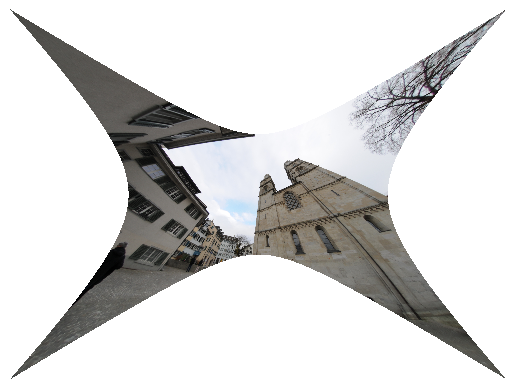}%
    \includegraphics[width=0.27\linewidth]{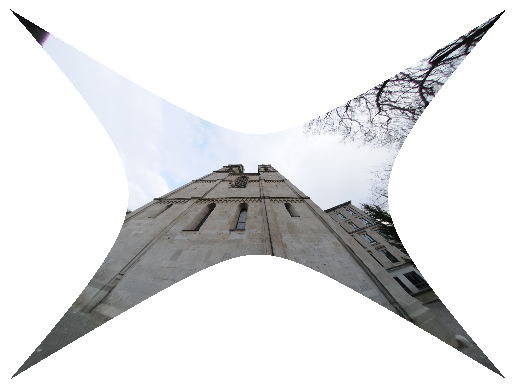}%
    \caption{%
        \emph{Grossmünster Church}.
        (Top) Inliers obtained using the proposed two-sided equal solver.
        (Bottom) Examples of rectified images obtained using the proposed algorithm.
        Note that the edges of the building appear straight, indicating a successful correction.
    }
    \label{fig:grossmunster_matches}
\end{figure*}

\paragraph{Evaluation on Grossmünster Church.}
Due to the lack of real-life homography benchmarks with radial distortion we use the Grossmünster Church dataset~\cite{eccv/2020/larsson_etal} which contains real images with fisheye distortion.
 In contrast to the HPatches dataset
distortion is present, and is not perfectly approximated by the one-parameter division model.
This dataset was originally used in RadialSfM~\cite{eccv/2020/larsson_etal}, and we use it to generate a 3D reconstruction of the scene, see~\cref{fig:front}.
In order to extract ground truth homographies, we use the acquired poses, and compute the relative pose~$\bR=\bR_2\bR_1^{\T}$ and relative translation~$\bt=\bt_2-\bR\bt_1$.
As suggested in~\cite{cvpr/2023/barath_etal}, we select only images with a single dominant plane---in our case we chose the wall containing the two church towers---and fitted a plane directly on the 3D model.
Among the images in the dataset, seven images were selected for which it was possible to extract six consecutive image pairs with more than \num{100} tentative image correspondences (including outliers), see~\cref{fig:grossmunster_gt}.
The ground truth homography was then computed as $\bH = \bK(\bR-\bt(\bn')^{\T}/d)\bK^{-1}$, where $\bK$ is the shared intrinsic parameters across all views, $\bn'=\bR_1\bn$ is the relative plane normal, and~$d$ the corresponding depth.

For the experiment, we extracted SIFT keypoints and applied Lowe's ratio test. The final correspondences for one image pair are shown in~\cref{fig:grossmunster_matches}.
Since radial distortion is present in both images, only the two-sided solvers are used in this comparison.
The homography error is computed as $\epsilon = \norm{\bH-\bH_{\textup{est}}}_{\textup{F}}$, where both $\bH$ and $\bH_{\textup{est}}$ are normalised to have unit Frobenius norm and positive determinant.

All solvers are configured to run LOMSAC \cite{bmvc/2012/lebeda_etal} with a minimum of \num{500} iterations and final non-linear least squares fitting is applied.
The average homography error is almost identical for all solvers; however, the execution times differ, see~\cref{tab:grossmunster}.

\begin{table}[ht!]
\centering
\caption{%
    \emph{Results on Grossmünster Church}.
    Mean error and mean execution time of the solvers in a LOMSAC framework.
    Best results are marked in bold (across both categories).
}
\label{tab:grossmunster}
    \begin{tblr}{colspec={XQ[r]Q[r]},rowsep=0ex,width=\linewidth}
        \toprule
        Solver &  Error & Exec. time \\
        \midrule
        \SetRow{ht=1.1\baselineskip}\SetCell[c=3]{c} Two-sided with independent $\lambda$ and $\lambda'$ \\
        \midrule
        \citet{cvpr/2015/kukelova_etal} (5pt)       & 0.039 & \qty{2657}{\milli\second}\\
        \citet{cvpr/2015/kukelova_etal} (6pt)       & 0.040 & \qty{330}{\milli\second}\\
        Proposed (5pt)                              & \textbf{0.038} & \qty{150}{\milli\second} \\
        \midrule
        \SetRow{ht=1.1\baselineskip}\SetCell[c=3]{c} Two-sided with $\lambda' = \lambda$ \\
        \midrule
        \citet{cvpr/2001/fitzgibbon} (5pt)          &  0.040 & \qty{233}{\milli\second}\\
        \citet{cvpr/2015/kukelova_etal} (5pt)${}^*$ &  0.040 & \qty{2053}{\milli\second}\\
        \citet{cvpr/2015/kukelova_etal} (6pt)${}^*$ &  0.039 & \qty{321}{\milli\second}\\
        Proposed (4.5pt)                            &  0.039 & \textbf{\qty{84}{\milli\second}} \\
        \bottomrule
    \end{tblr}
\end{table}

\section{Conclusion}\label{sec:conclusion}
In this paper, we have revisited the fundamental problem of simultaneously estimating a homography and radial distortion.
Basing our analysis on a classical closed-form solution instead of the DLT, we have taken a unified approach to derive three new, fast, stable and accurate solvers for the possible cases of radially distorted homographies, \ie \begin{enumerate*}[label={(\roman*)}] \item the one-sided case, \item the two-sided equal case, and \item the two-sided independent case. \end{enumerate*}
We have evaluated our proposed solvers against existing state-of-the-art solvers for each of the three cases, and our new solvers either outperform or compare favourably to the state-of-the-art in these evaluations.

%% file: sec/2_acknowledgements.tex
\section*{Acknowledgements}
This work was supported by the \emph{Wallenberg Artificial
Intelligence, Autonomous Systems and Software Program}
(WASP), funded by the Knut and Alice Wallenberg Foundation, as well as by the strategic research environment ELLIIT, funded by the Swedish government.